\documentclass[fleqn,10pt]{wlscirep}
\usepackage[utf8]{inputenc}
\usepackage[T1]{fontenc}
\usepackage{listings}
\usepackage{caption}
\usepackage{subcaption}

\usepackage{tikz-dependency}
\usepackage[autostyle]{csquotes}
\usepackage{placeins}

\makeatletter
\renewcommand*{\@fnsymbol}[1]{\ifcase#1\or*\else\@arabic{\numexpr#1-1\relax}\fi}
\makeatother

\title{Wiki-Quantities and Wiki-Measurements: Datasets of Quantities and their Measurement Context from Wikipedia}

\author[1,2,*]{Jan Göpfert}
\author[1]{Patrick Kuckertz}
\author[1]{Jann M. Weinand}
\author[1,2]{Detlef Stolten}
\affil[1]{Forschungszentrum J{\"u}lich GmbH, Institute of Climate and Energy Systems, J{\"u}lich Systems Analysis, 52425 Jülich, Germany}
\affil[2]{RWTH Aachen University, Chair for Fuel Cells, Faculty of Mechanical Engineering, 52062 Aachen, Germany}

\affil[*]{corresponding author(s): Jan Göpfert (j.goepfert@fz-juelich.de)}

\begin{abstract}
To cope with the large number of publications, more and more researchers are automatically extracting data of interest using natural language processing methods based on supervised learning. Much data, especially in the natural and engineering sciences, is quantitative, but there is a lack of datasets for identifying quantities and their context in text. To address this issue, we present two large datasets based on Wikipedia and Wikidata: \emph{Wiki-Quantities} is a dataset consisting of over 1.2 million annotated quantities in the English-language Wikipedia. \emph{Wiki-Measurements} is a dataset of 38\,738 annotated quantities in the English-language Wikipedia along with their respective measured entity, property, and optional qualifiers. Manual validation of 100 samples each of Wiki-Quantities and Wiki-Measurements found 100\% and 84-94\% correct, respectively. The datasets can be used in pipeline approaches to measurement extraction, where quantities are first identified and then their measurement context. To allow reproduction of this work using newer or different versions of Wikipedia and Wikidata, we publish the \href{https://github.com/FZJ-IEK3-VSA/wiki-measurements}{code} used to create the datasets along with the \href{https://doi.org/10.5281/zenodo.14858280}{data}.
\end{abstract}

\begin{document}

\flushbottom
\maketitle

\thispagestyle{empty}

\section*{Background \& Summary}

As the volume of publications continues to grow, scientists are increasingly challenged to keep track of their research field.
Information extraction is used to transform aspects of publications into structured information to enable large-scale literature analyses\cite{courtAutogeneratedMaterialsDatabase2018,foppianoSuperMatConstructionLinked2021}. Much of the data, especially in the natural and engineering sciences, is quantitative, and therefore measurement extraction is of great importance. While information extraction traditionally focuses on identifying mentions of named entities and the relationships between them, measurement extraction specifically deals with the identification of quantities and their measurement context, such as their measured properties and entities~\cite{gopfertMeasurementExtractionNatural2022}. 
In a previous review, we show that measurement extraction has received little attention in the history of information extraction, which is reflected in a lack of large, high-quality datasets for extracting quantities and their measurement context~\cite{gopfertMeasurementExtractionNatural2022}. As measurement extraction is typically approached with supervised learning, the lack of annotated data limits the performance of measurement extraction systems. The task of measurement extraction is typically approached in a pipeline manner, where 1) quantities are identified before 2) their individual measurement context is extracted~\cite{gopfertMeasurementExtractionNatural2022}. To support the development and evaluation of measurement extraction systems, we present two datasets that correspond to the two tasks:
\begin{itemize}
    \item \textbf{Wiki-Quantities}, a dataset for identifying quantities, and
    \item \textbf{Wiki-Measurements}, a dataset for extracting measurement context for given quantities.
\end{itemize}
    
\begin{figure}[h]
    \centering
    \includegraphics[width=0.992\linewidth]{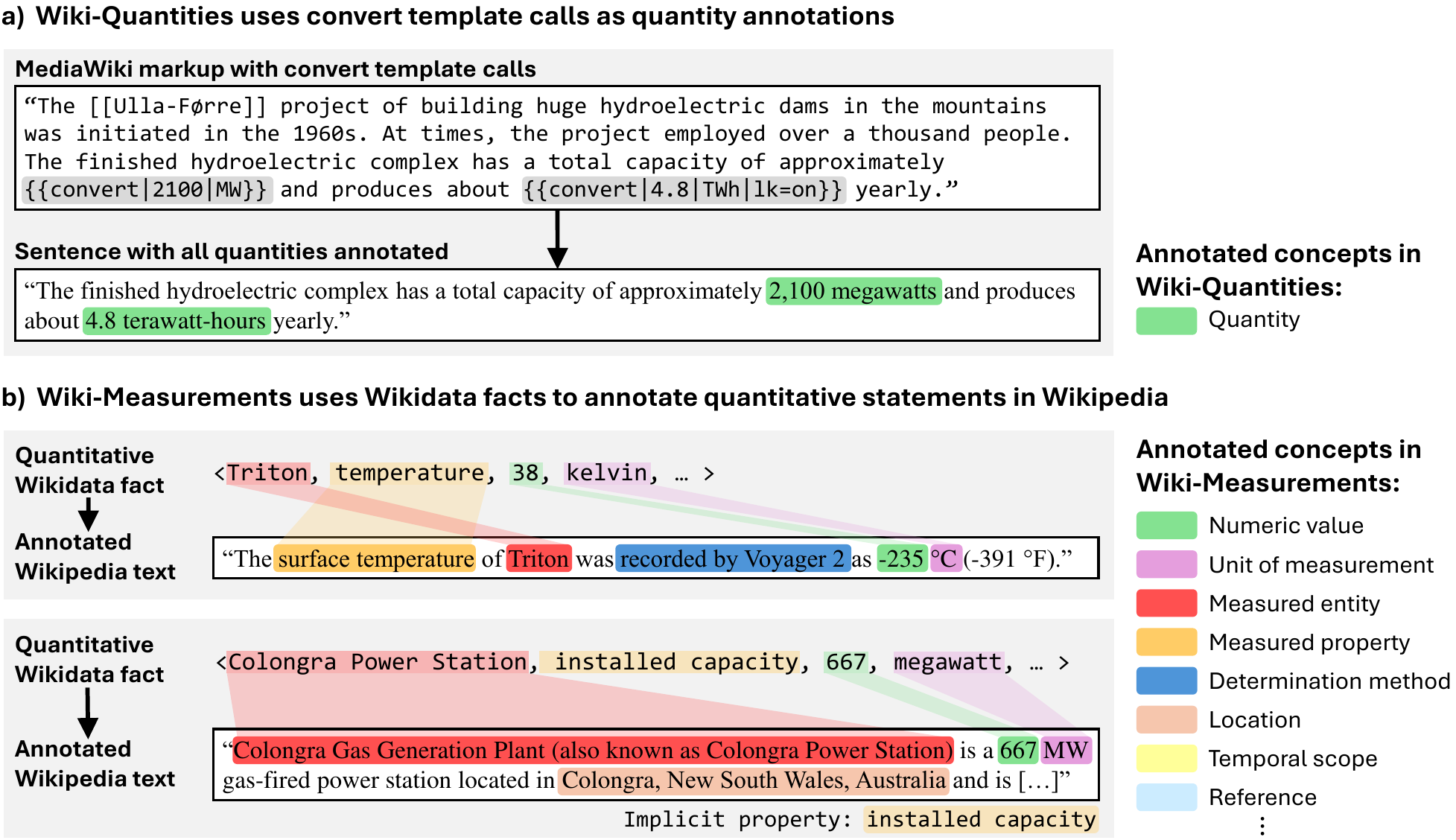}
    \caption[Graphical Abstract]{Schematic illustration of the methodology used to create the datasets: a) Wiki-Quantities uses convert template calls as quantity annotations. b) Wiki-Measurements aligns Wikidata facts with text from their respective Wikipedia page. Both implicit and explicit properties are supported. The text examples are from Wikipedia\protect\footnotemark.}
    \label{fig:graphical_abstract}
\end{figure}

\noindent In Wiki-Quantities, each example is a sentence in which all quantity spans are annotated, such as the example in 
Figure~\hyperref[fig:graphical_abstract]{1a}. In contrast, Wiki-Measurements consists of sentences in which the measured entity and property are annotated for a single quantity. The measured property may be given implicitly or explicitly (see Figure~\hyperref[fig:graphical_abstract]{1b}). 
Qualifiers such as temporal or spatial scopes or measurement methods and quantity modifiers such as `approximately' or `at least' are considered optional. Therefore, qualifier and quantity modifier annotations are only sparsely included. In Wiki-Measurements, quantity annotations are divided into value and unit annotations.
In the following, we give an overview of related datasets, all of which either have a different scope, are considerably smaller and/or are not published in a FAIR~\cite{wilkinsonFAIRGuidingPrinciples2016} manner. \\

\subsection*{Datasets of annotated numerals and quantities}
\footnotetext{Wikipedia pages used as examples: \\ \url{https://en.wikipedia.org/w/index.php?title=Suldal&oldid=1267112000}, Accessed on: 2025-03-04 \\ \url{https://simple.wikipedia.org/w/index.php?title=Triton_(moon)&oldid=9749756}, Accessed on: 2025-03-04 \\ \url{https://en.wikipedia.org/w/index.php?title=Colongra_Power_Station&oldid=1222955505}, Accessed on: 2025-03-04}
Numerical concepts such as percentages, monetary expressions and quantities have long been part of tag sets in named entity recognition~(NER) datasets~\cite{chinchorOverviewMUC71998,grishmanDesignMUC6Evaluation1995,weischedelralphOntoNotesRelease502013}. Today, several datasets are specifically targeted at numerical information. Many of these, such as \emph{Numeracy-600K}\cite{chenNumeracy600KLearningNumeracy2019}, \emph{NumerSense}\cite{linBirdsHaveFour2020}, and others~\cite{spithourakisNumeracyLanguageModels2018,berg-kirkpatrickEmpiricalInvestigationContextualized2020}, contain annotations of numerals without units of measurement. However, many numerals are not part of a quantity (e.g., 15 in `iPhone 15'). The applicability of these datasets for measurement extraction is further limited by considering only numerals that are written out (i.e. `ten' instead of `10')~\cite{linBirdsHaveFour2020}, omitting numerals outside a range~\cite{berg-kirkpatrickEmpiricalInvestigationContextualized2020}, not including mathematical symbols and therefore ignoring signs~\cite{berg-kirkpatrickEmpiricalInvestigationContextualized2020}, or splitting fractions into multiple numerals~\cite{spithourakisNumeracyLanguageModels2018}. 

In contrast, \emph{Grobid-quantities}~\cite{foppianoAutomaticIdentificationNormalisation2019} distinguishes between quantities and other numerals. In addition, quantities are classified into different types (e.g., list, range, least, most, etc.). Both the numeric value and the unit of a quantity are annotated. The numeric value is further subdivided into exponent, power, base, time expressions, and alphabetic numbers. The unit is further divided into prefix, base, and power. However, the dataset consists of only 35 labelled documents\footnote{\url{https://github.com/kermitt2/grobid-quantities/tree/master/resources/dataset/quantities},\\ Accessed on: 2024-08-29}.
Similarly, \emph{MeasEval}~\cite{harperSemEval2021TaskMeasEval2021} distinguishes between quantities and other numerals, annotates both the value and the unit, and classifies a quantity into different types (e.g., range, list, approx., mean, etc.).
\emph{CQE}~\cite{almasianCQEComprehensiveQuantity2023} provides a test set that lists all normalized quantities per sentence, but does not indicate their position. Other datasets distinguish between numeric values and units, but only annotate certain quantities~\cite{chagantyHowMuch1312016}.

Most similar to our work, \emph{Wiki-Convert}~\cite{thawaniNumeracyEnhancesLiteracy2021} contains 924\,479 sentences from the English Wikipedia with annotated numbers. The aforementioned rule-based approaches do not distinguish between numerals as part of a quantity, nominal numbers (e.g., zip codes) and ordinals. In Wiki-Convert, annotations are drawn from \{\{convert\}\} template calls, so numbers are more likely used as a quantity. Sentences with multiple number annotations and \{\{convert\}\} templates involving a unit outside the 30 most common units are skipped. It remains unclear how the  \{\{convert\}\} template calls were processed into quantity strings, as the code used to create the dataset is not published. Wiki-quantities is a re-creation of Wiki-Convert that removes some of its limitations and is published in a FAIR manner.

\subsection*{Datasets of annotated quantities and their context}
Going beyond mere numerals, several datasets such as \emph{NumER}~\cite{julavanichNumERFineGrainedNumeral2021}, \emph{FinNum 1.0}~\cite{chenNumeralUnderstandingFinancial2018}, \emph{FiNER-139}~\cite{loukasFiNERFinancialNumeric2022a}, \emph{NumClaim}~\cite{chenNumClaimInvestorFinegrained2020}, and \emph{ECNum}~\cite{chenDistillingNumeralInformation2021}
additionally state the numeral type (e.g., age, population, or money), yet without including units in the annotations. Other datasets focus on specific entities and quantities: \emph{SuperMat Corpus}~\cite{foppianoProposalAutomaticExtraction2019} focuses on mentions of superconducting materials and their critical temperature; the \emph{SOFC-Exp Corpus}~\cite{friedrichSOFCExpCorpusNeural2020} focuses on materials, quantities, and devices describing experiments related to solid oxide fuel cells; and the \emph{Materials Science Procedural Text Corpus}~\cite{mysoreMaterialsScienceProcedural2019} focuses on materials synthesis procedures.
Other datasets, such as \emph{BioNumQA}~\cite{wuBioNumQABERTAnsweringBiomedical2021}, \emph{NQuAD}~\cite{chenNQuAD700002021}, \emph{FinQA}~\cite{chenFinQADatasetNumerical2021}, \emph{DROP}~\cite{duaDROPReadingComprehension2019a}, or datasets of math word problems require models to perform numerical reasoning and mathematical operations, but do not focus on detailed annotation of quantities and their context. 

\emph{MeasEval}~\cite{harperSemEval2021TaskMeasEval2021}, already mentioned as a dataset of annotated quantities, additionally provides annotations for their measured entities and properties as well as qualifiers. Qualifiers describe further context that is relevant to the particular quantification. However, the dataset consists of only 428 labeled paragraphs, which limits the performance of machine learning models trained on it~\cite{lathiffCLaCnpSemEval2021Task2021}. 
\emph{Chaudron}~\cite{subercazeChaudronExtendingDBpedia2017} is a dataset of measurements extracted from Wikipedia infoboxes. 
Methodologically related to our approach, Wikipedia infoboxes have been used to heuristically generate training data from the articles in which they are embedded in~\cite{hoffmannLearning5000Relational2010}. Similarly, 
\emph{LUCHS}~\cite{hoffmannLearning5000Relational2010} is a system for extracting triples from text that is distantly supervised by aligning facts from Wikipedia infoboxes with the corresponding article. Most similar to our work, Chia et al.~\cite{chiaDatasetHyperRelationalExtraction2022} and Chen et al.~\cite{chenTimelinebasedSentenceDecomposition2024} align Wikidata~\cite{vrandecicWikidataFreeCollaborative2014} facts with Wikipedia articles, but they do not focus on quantitative facts.

\section*{Methods}
\label{sec:methods}
Many systems for measurement extraction first identify quantities before extracting their measurement context~\cite{gopfertMeasurementExtractionNatural2022}. 
This pipeline approach not only results in powerful systems but also facilitates the creation of datasets. Whereas a dataset for quantity span identification must provide annotations for all quantities that occur in a text, a dataset for measurement context extraction may provide sparse annotations for quantities, but for given quantities must provide complete annotations of their measurement context. In the following subsections, we describe the largely automated creation of large datasets for quantity and measurement context extraction, which is made feasible due to their different requirements. To improve the reproducibility of this work, all code is published open-source and accompanied by a Snakemake workflow that formalizes all the steps involved in generating the raw datasets. In addition, Wiki-Quantities and Wiki-Measurements have been deduplicated, balanced, and partially manually curated.

\subsection*{Wiki-Quantities}
\label{sec:quantity_dataset}
Wikipedia authors use templates to reuse frequent content or to automate certain functions when editing Wikipedia articles. Similar to function calls in programming, templates are invoked by specifying their name and arguments in double curly braces directly within the MediaWiki markup, which can be thought of as the source code of a Wikipedia article. When rendering the HTML representation of a Wikipedia article, MediaWiki replaces template calls with their returned values. One such template is the \{\{convert\}\} template which is used to automatically convert quantities from one unit to another. For example, the string
\noindent\verb@{{convert|11|m/s|km/h|abbr=on}}@\footnote{\url{https://en.wikipedia.org/w/index.php?title=Barn_swallow&action=edit}, Accessed on: 2024-08-29}
will be rendered as ``{11 m/s (40 km/h)}''\footnote{\url{https://en.wikipedia.org/wiki/Barn_swallow}, Accessed on: 2024-08-29}.
More complex calls are also supported. For example,
\noindent\verb@{{convert|60-62.5|m|ft+royal cubit|...}}@\footnote{\url{https://en.wikipedia.org/w/index.php?title=Pyramid_of_Djoser&action=edit}, Accessed on: 2024-08-29},
is displayed as ``{60–62.5 m (197–205 ft; 115–119 cu)}''\footnote{\url{https://en.wikipedia.org/wiki/Pyramid_of_Djoser}, Accessed on: 2024-08-29}.
To create the quantity dataset, we take \{\{convert\}\} template calls in the MediaWiki markup as quantity annotations, as previously described in Thawani et al.\cite{thawaniNumeracyEnhancesLiteracy2021}.

\subsubsection*{Parsing MediaWiki markup and \{\{convert\}\} template expansion}
For each article in the English Wikipedia and Simple English Wikipedia\footnote{Using database dumps from Q3 2022}---written with simplified vocabulary and grammar---we parse the MediaWiki markup into readable text, remove infoboxes, sidebars, references, and comments, and expand all \{\{convert\}\} templates, that is, replace the template calls with the results returned by the corresponding \{\{convert\}\} module. 
For this, we isolated the \{\{convert\}\} module written in the Lua language from MediaWiki, in which it is interwoven, and invoke it directly from within Python. The returned strings are taken as the quantity annotations. Any template calls that reference Wikidata are ignored. We customized the \{\{convert\}\} module to return three versions of a quantity string: the quantity in input units, the quantity in output units, and both combined into one string. We randomly select one of these versions, giving the input and output quantities the same probability of 47.5\%. The combined string is only given a probability of 5\% because it occurs less frequently in other corpora.

\subsubsection*{Post-processing}
We split the text into sentences using spaCy~\cite{montaniExplosionSpaCyNew2022} and keep only those that include at least one quantity annotation. To ensure these do not include further quantities that were not specified using the \{\{convert\}\} template and therefore not annotated, we filter out sentences where digits, number words, and so forth appear outside the quantity annotations. However, based on the NER annotations\footnote{Whitelisted NER tags: DATE, EVENT, FAC, GPE, LANGUAGE, LAW, LOC, NORP, ORG, PERSON, PRODUCT, TIME, WORK\_OF\_ART} provided by spaCy, we allow products, and so forth to contain numerals. We drop sentences containing ``per'' outside of quantity annotations to filter out incomplete quantities such as ``{[2 million barrels]}\textsubscript{Quantity} per day'', as well as sentences shorter than 10 characters.

To provide a training incentive not to confuse citation and reference spans (e.g., `[1]', `\textsuperscript{2-3}', `Fig. IV' or `Table A1') with quantities, we add citation and reference spans to the end of sentences. We sample a diverse and realistic set of citation and reference spans from the S2ORC corpus~\cite{loS2ORCSemanticScholar2020a}---a large dataset of scientific articles that has been parsed into a machine-readable format. 
We add phrases like "cf." or "as shown in" to some of them and enclose them in parentheses () or brackets [], where the probability of using parentheses or brackets depends on the type of citation or reference. For example, citations such as "Einstein et al., 2024" or references to figures and tables are less likely to be enclosed in brackets than numerical references. We place all figure and table references (e.g., ``Fig. 1'') before the full stop and bibliographic references (e.g., ``[1]'') after the full stop with a 5\% probability.

Examples that would be longer than 512 word-piece tokens using the Hugging~Face implementation of the RoBERTa~\cite{liuRoBERTaRobustlyOptimized2019} tokenizer are split into multiple examples. We manually curated a subset of the data and corrected the entire dataset for common errors. This included removing empty parentheses (``( ; )''), deleting repetitions of units (``3.1 metres metres''), removing `-high', `-long', etc. from the quantity span (``3.1-meter-high''), and expanding the quantity annotation where necessary (``'1 meter' square'', ``'3.1 in' for every second'', ``'38 °C'±1''). The result is a high-quality dataset of sentences with annotated quantities.

\subsubsection*{Deduplication and balancing}
However, the raw dataset is imbalanced with respect to units. Expressions for meter, kilometer, feet, inch, miles, acres, kilogram, $^{\circ}$C, km/h, mph, and so forth, are predominant. Accordingly, the dataset is dominated by the description of lengths, heights, distances, weights, and so forth. Additionally, many of the most frequent words either describe properties directly (such as `long', `length', `highest', `speed', or `altitudes') or limit the number of properties to expect by setting a topic (such as `river', `lake', `town', `mountain', or `highway'). In addition, the dataset contains duplicates and many similar examples, which we hypothesize are the result of bots and templates being used to create Wikipedia pages for certain frequent but repetitive article classes, such as road networks, rivers, mountains, municipalities, lighthouses, and so on. 

We remove examples that exceed a certain similarity threshold in their normalized\footnote{Normalization includes converting all characters to lowercase, replacing all quantity spans with a special character, removing leading and trailing whitespace as well as references at sentence ends, replacing all directions with `north', all years with `2023', all months with `January, and common terms referring to settlements (e.g., city or village) with `region'.} form. As a measure of similarity, we use the token-based Levenshtein distance~\cite{levenshteinBinaryCodesCapable1966}. Tokenization is done by splitting on whitespace. If the absolute value of the edit distance is greater than 35\% of the number of tokens, the example is considered unique, otherwise it is removed from the dataset. In this way, we removed 37\,530 duplicate examples and a further 233\,771 examples based on the edit distances, which equates to a 32.6\% reduction in the number of examples.
561\,393 examples remain. We refer to the original dataset as Wiki-Quantities (raw) and to this scaled-down version as Wiki-Quantities (large). Because Wiki-Quantities is orders of magnitude larger than other quantity span identification datasets, reducing its size helps avoid overfitting to it in scenarios where a model is trained on a mixture of quantity span identification datasets. 

To balance the dataset with respect to the units, properties, and topics, we remove examples based on unit and token counts. 
For this, we compile a list of the most frequent tokens, remove neutral tokens (e.g., quantity modifiers, such as `approx.', and stopwords), and randomly remove examples that contain any of them. The probability increases with the frequency of the token and is given by $1 - t_{thr} / (t_{i} + t_{thr})$, where $t$ is the number of occurrences of token i in the dataset and $t_{thr}$ a parameter to control the degree of filtering. Hereinafter, we limit the number of examples containing a quantity span ending on a specific unit in the dataset to a threshold count of $u_{thr}$. Both $u_{thr}$ and $t_{thr}$ control the degree of filtering. Depending on their value, we get datasets of different sizes. In addition to Wiki-Quantities (raw) and Wiki-Quantities (large), we publish a small and tiny variant corresponding to a $(u_{thr},\,t_{thr})$ of $(1200, 1000)$ and $(300, 1000)$, respectively (see Table~\ref{tab:wiki_quantities_stats}). Table~\ref{tab:wiki_quantities_unit_freqs} compares the occurrence frequencies of units in the Wiki-Quantities variants. As to be expected, Wiki-Quantities (small) and (tiny) are more balanced with respect to units than the other variants.

\subsection*{Wiki-Measurements}
Distant supervision is an approach to deal with data scarcity in relation extraction tasks by labeling relations between entities based on known triples in a knowledge base~\cite{mintzDistantSupervisionRelation2009a}. The creation of the measurement context dataset is based on distant supervision, where matches to a knowledge base are treated as (somewhat noisy) ground truth. Our knowledge base consists of quantitative facts from \emph{Wikidata}~\cite{vrandecicWikidataFreeCollaborative2014}, an openly licensed, collaboratively developed knowledge graph that is part of the Wikimedia Foundation. 
At the time of writing, Wikidata contains information on more than 113 million items\footnote{\url{https://www.wikidata.org/wiki/Wikidata:Statistics}, Accessed on: 2024-08-29}, including many quantitative statements such as \verb@<Airbus A380, wingspan, 79.8, metre>@\footnote{\url{https://www.wikidata.org/wiki/Q5830}, Accessed on: 2024-08-29}. Wikidata publishes RDF dumps that can be queried via its SPARQL endpoint\footnote{\url{https://query.wikidata.org/}}. RDF is a data model standardized by the World Wide Web Consortium, in which information is represented in triples containing a subject, a predicate and an object. We query Wikidata for quantitative statements and align these with associated Wikipedia articles to create text chunks with annotated quantity, property, entity, and optionally qualifiers.

\subsubsection*{Obtaining quantitative statements from Wikidata}
First, we query Wikidata for statements whose subject is an item that has an associated article in the English or Simple English Wikipedia, whose predicate is of type `quantity', and whose object is composed of a numeric value and a unit. Counts, that is, quantities without a unit, are also considered, as they have the unit `1' in Wikidata. 
Additionally, we get the lower and upper bounds of the numeric value, if specified. We also query for qualifiers that constrain quantitative statements, for example, in terms of measurement method, date, or precision. Qualifiers may be qualitative or quantitative. In the latter case, we again retrieve the numeric value, the unit of measurement, and the lower and upper bounds. For temporal qualifiers, we get the time and its precision (such as year or day). When we hit the timeout of Wikidata's public SPARQL endpoint, we use the results obtained up to that point, as this naturally balances the dataset and we can do without loading the Wikidata dump into a local triplestore. We retrieve 636 different predicates of type `quantity'. Limiting the number of results per predicate to 250\,000, we retrieve 3\,018\,117 and 622\,295 quantitative statements for the English and Simple English Wikipedia, respectively. The queries are given in Query~\ref{lst:property_query} and~\ref{lst:quantitative_statement_query}.

\subsubsection*{Aligning Wikidata facts and their associated Wikipedia articles}
Next, we align the quantitative statements obtained from Wikidata with text chunks of their associated Wikipedia articles. Analogous to the creation of Wiki-Quantities, we parse the MediaWiki markup into plain text, perform sentence segmentation, and expand \{\{convert\}\} template calls. To be considered a valid example of the dataset, a sentence must contain the numeric value as well as the label or any alternative label of the measured entity, property, and unit. 
As properties are often given implicitly, a matching property span is considered optional if other evidence provides sufficient confidence that the quantitative statement is correctly aligned with the Wikipedia page.\footnote{The numeric value must be infrequent (that is, not 1-10) and must match exactly. Additionally, a matching entity is required.} In this case, we use the Wikidata statement's property label as the annotation. 

To match the numeric value, we attempt to identify all numbers using a combination of regular expressions, gazetteers, and rules on ``\verb@CD@'' and ``\verb@like_num@'' tags provided by spaCy. We filter out ordinals, dates, and numbers preceded by ``per''. We perform \textbf{approximate matching} of numeric values, that is, if bounds are given, the numeric value must lie within the corresponding interval, otherwise the numeric value is allowed to deviate by a mean absolute percentage error~(MAPE) of at most 3\% in order to match.
Additionally, we perform \textbf{unit conversions} to increase recall. The same quantity can be expressed in different units of measurement, hence we convert the quantities of the Wikidata  statements into various units when matching them. For performance reasons, we only consider unit conversions that have a conversion factor in the interval (10\textsuperscript{-3}; 10\textsuperscript{3}). For counts, we take succeeding noun phrases as unit (e.g., `seat' in ``a 2,564-seat concert hall''). A number of rules improve the alignment precision: Units may not be preceded by strings such as ``of which''. If there are multiple matches for a value in the same sentence, but only one is an exact match that is not a count (that is, without a unit), that one is taken. Otherwise, none is taken because there is not enough evidence to decide which value is correct. If a single unit matches that is not adjacent to the target value, it is not allowed to be adjacent to any other value, nor is the target value allowed to be adjacent to any other unit. If multiple units match, but only one is adjacent to the value, that one is taken. Otherwise, the example is discarded. 

The labels of units, measured entities, and properties are matched based on their lemmas. Depending on the concept type, annotation spans are expanded based on dependency parsing, part-of-speech tags, and/or NER annotations. %\verb@amod@ or \verb@compound@ dependency tags, the \verb@PROPN@ POS tag, and/or NER annotations. 
As width, height, depth, and thickness are common properties that are often used interchangeably, we use rules to prevent them from being mistakenly associated. If multiple spans match an entity or property, either the one that is in the same subclause as the other annotations or the one that gives the shortest overall path between annotations based on dependency parsing is chosen. 

Additionally, we try to match \textbf{qualifiers} that we consider optional. We distinguish the following qualifiers: point in time (Wikidata ID \verb@P585@\footnote{P585 etc. are IDs of properties in Wikidata (e.g., P585 is ``point in time'' \url{https://www.wikidata.org/wiki/Property:P585}, Accessed on: 2024-08-29}), start time (\verb@P580@), end time (\verb@P582@), location (\verb@P276@), coordinate location (\verb@P625@), applies to part (\verb@P518@), in scope of (\verb@P642@), criterion used (\verb@P1013@), determination method (\verb@P459@), and according to (\verb@P3680@). All other qualifiers are placed under a generic label (``other qualifiers''). 
Geographic coordinates and temporal scopes are matched using special rules. We do not match rankings (\verb@P1352@), which are qualifiers used to rate the quality of conflicting statements within Wikidata.
For values less than ten, we discard examples without qualifiers because small integers are common and therefore more likely to be mistakenly matched. 

After obtaining all examples for a Wikipedia page, we remove duplicate examples and \textbf{resolve contradictions}. Sometimes different facts lead to conflicting annotations for a sentence, or the same training example is created multiple times, because there are multiple similar coexisting statements about a property (e.g., the area of Alabama is stated to be 135\,765, 134\,000$\pm$500 or 131\,365$\pm$0.5 square kilometres\footnote{\url{https://www.wikidata.org/wiki/Q173}, Accessed on: 2024-08-29}). Comparing a pair of examples for the same sentence at a time, the following filter rules are applied: If two examples are identical, we drop one of them. If both match, but one has additional qualifiers, the one with fewer qualifiers is dropped. If both have qualifiers defined and they contradict each other, both are dropped. If two examples share the same value but differ in their entity, property or unit annotation, or if the entity, property, unit and qualifier annotations match but the values do not, both examples are dropped.

\subsubsection*{Deduplication and balancing}
We remove duplicates and refer to the dataset as Wiki-Measurements (large). Because the raw dataset differs from the large dataset by only 72 duplicates, the raw dataset is not published. A variant with near-duplicates removed is referred to as Wiki-Measurements (small). In addition, we publish variants where the quantitative facts are more strictly aligned with the text (large-strict, small-strict). In this case, we do not accept examples where the property cannot be matched but a rare numeric value is exactly matched, where the entity can only be matched using co-reference resolution, where a single annotation is selected based on shortest path, or where subclause membership is used. In addition, the threshold for small rounded values is lowered. Furthermore, we publish variants with preceding and following sentences added around the annotated sentence (small-context, small-context-strict). The published dataset is accompanied by additional information for each example, detailing the Wikidata fact, the weak acceptance reasons if any, and the context before and after the annotated sentence, allowing anyone to create further variants.

For further details, the reader is referred to the published source code. The result is a dataset of sentences, each annotated with a single quantity and its measurement context. 

\section*{Data Records}
\FloatBarrier
\label{sec:data_records}
Wiki-Quantities and Wiki-Measurements are published at \url{https://doi.org/10.5281/zenodo.14858280}.

We publish different variants of \textbf{Wiki-Quantities}: The raw variant, the large variant where duplicates and similar examples are removed, and two further filtered variants, referred to as small and tiny, which are more balanced in terms of units and topics. In addition, we publish the data in a raw format and pre-processed for IOB sequence labeling.
For the different variants of Wiki-Quantities, Table~\ref{tab:wiki_quantities_stats} lists the number of quantity annotations, examples, and Wikipedia pages they originate from. In total, the raw, large, small and tiny variants contain 832\,660, 561\,393, 44\,729, and 20\,148 examples, respectively. 289 examples have been manually curated and are included in all variants. The raw dataset contains 1\,227\,626 quantity annotations in 434\,860 different Wikipedia pages. 
Table~\ref{tab:wiki_quantities_unit_freqs} lists the 50 most frequent units for the different Wiki-Quantities variants\footnote{Note that unlike Wiki-Measurements, Wiki-Quantities does not include separate value and unit annotations. Hence, the units are determined by rule-based parsing of the quantity span.}. Different expressions for units of length (mi, km, etc.), area (acres, km2, etc.), velocity (km/h, mph, etc.), temperature (°F, °C), mass (kg, lb, etc.), volume (m3, L, etc.), and power (kw, hp, etc.) predominate. The approach to balancing the dataset is effective in leveling the unit shares. Table~\ref{tab:wiki_quantities_language_stats} compares how many quantity annotations and examples are drawn from the English language and Simple English language Wikipedia, respectively. With only 1.2\% of all examples stemming from the Simple English Wikipedia, the standard English variant clearly predominates. In Wiki-Measurements (raw), 4.4\% of all examples are drawn from the Simple English Wikipedia.

Like Wiki-Quantities, we publish \textbf{Wiki-Measurements} in several variants, as well as in a raw format and pre-processed for SQuAD-style generative question-answering. Wiki-Measurements (large) contains 38\,738 examples; the small variant contains 21\,854 examples. The number of annotations per concept type, as well as the number of examples and pages from which they originate, is given in Table~\ref{tab:wiki_measurements_stats}. Each example has an entity, a property, a value, and, if applicable, a unit annotation.  Annotations for quantity modifiers and qualifiers are considered optional. Thus, quantitative statements may have additional qualifiers that are not annotated. 214 examples were manually curated to ensure complete annotations of the spatio-temporal scope and other qualifiers. The most frequent entities, properties, and units in the deduplicated (large) and balanced (small) variant of Wiki-Measurement are given in Table~\ref{tab:wiki_measurements_freq_deduplicated} and Table~\ref{tab:wiki_measurements_freq_balanced}, respectively.

\FloatBarrier
\section*{Technical Validation}
The quality of the datasets was primarily ensured by the approaches used to generate them and their elaborate, iteratively defined filter rules. 

\paragraph*{Wiki-Quantites.}
Wiki-Quantities is created based on \{\{convert\}\} template calls. The \{\{convert\}\} templates are used to render quantities. Therefore, it is generally correct to consider their output as quantity span annotations. To be sure that all quantities in a sentence have been annotated, filter rules apply if there are numbers outside the template calls. Randomly sampling 100 examples of Wiki-Quantities (large) and manually validating them gave \textbf{100\% precision and recall}. Nevertheless, the dataset contains incorrect examples. Therefore, we train a quantity span identification model on Wiki-Quantities (small) and other datasets, and compare the model outputs with the annotations for Wiki-Quantities (small), adjusting the annotations where necessary. We transfer the changes to Wiki-Quantities (tiny) and (large). As a superset, this leaves Wiki-Quantities (large) with more undetected mistakes than the smaller variants.

\paragraph{Wiki-Measurements.}
Wiki-Measurements is created based on distant supervision which naturally yields noisy labels. However, we minimize noisy labels by only matching facts from the knowledge base (Wikidata) with texts related to the subject of the fact (its Wikipedia article). Furthermore, aligning the predicate (measured property) in addition to the subject and object further increases accuracy, but at the cost of neglecting expressions where the predicate is implicit. Therefore, implicit properties are considered if the value, unit, and entity matches provide sufficient evidence for a correct alignment. In addition, many more filtering rules ensure the correct alignment of facts with texts. However, distant supervision with quantitative facts is prone to several errors which we also observe:

\begin{itemize}
    \item Often a sentence contains all the elements of a fact, but it is about a different thing that happens to share some of those elements. 
    \item Numeric values occur more frequently and in more contexts than traditional entities in relation extraction, such as the name of a particular company or person~\cite{madaanNumericalRelationExtraction2016}. 
    \item Distant supervision with quantitative facts requires approximate matching due to different levels of rounding~\cite{madaanNumericalRelationExtraction2016,vlachosIdentificationVerificationSimple2015,intxaurrondoDiamondsRoughEvent2015}, which introduces false positives. 
    \item Quantitative facts can be easily re-scoped (e.g., ``literacy rate of \emph{rural} India''~\cite{madaanNumericalRelationExtraction2016}). Therefore, we extend matching entity and property spans to include modifiers such as `rural', `minimum', etc. In some cases, an entity or property is extended incorrectly.     
    \item Time series in Wikidata (e.g., the growing population of a city) have many potential matches, making false positives more likely. 
\end{itemize}

Randomly sampling 100 examples of Wiki-Measurements (large) and manually validating quadruples of <entity, property, value, and unit>, yielded \textbf{84\% accuracy when judged strictly and 94\%} accuracy when essentially correct but not perfect examples were considered correct. Of these ten examples considered essentially correct, four are missing ``number of'' in a property span for a count (e.g., in \emph{``In 2019, {[Reipertswiller]}\textsubscript{Entity} had {[850]}\textsubscript{Value} {[inhabitants]}\textsubscript{Unit}''}, the property should be `number of inhabitants', not `inhabitants'), three have an entity span that is too short (e.g.,  \emph{``{[Dalesbred]}\textsubscript{Entity} ewes weigh...''}), two do not include an alternative name in the entity span (e.g., \emph{``{[Disentis]}\textsubscript{Entity}/Must\'er has an area...''}, and one is imprecise by describing a weightlifter's weight class as actual mass. 

Of the six incorrect examples, three contain the entity, property, and quantity of the Wikidata fact, but actually describe a different entity or property. For example, given the fact \emph{<Whittier, elevation above sea level, 13±1 metre>}, the sentence is incorrectly annotated as \emph{``The tsunami that hit {[Whittier]}\textsubscript{Entity} reached a {[height]}\textsubscript{Property} of 13 m ({[43]}\textsubscript{Value} {[ft]}\textsubscript{Unit}) and killed 13 people.''}. Coincidentally, the height of the tsunami is about the same as the city's elevation above sea level, colloquially height.
In another example, a sculpture and the person it represents are confused, because they have the same name. Given the fact \emph{<William Penn, height, 37 foot>} the sentence is incorrectly annotated as \emph{``Due to {[William Penn]}\textsubscript{Entity} being a slave owner, columnist Stu Bykofsky for "The Philadelphia Inquirer" sardonically wrote: "We can\'t abide {[37]}\textsubscript{Value} {[feet]}\textsubscript{Unit} of him towering over the city. (implicit property: height)''}. In one example, the property span is incorrectly extended (\emph{``{[Rolleston]}\textsubscript{Entity} is described by Statistics New Zealand as a {[medium urban area]}\textsubscript{Property}, and covers {[8.53]}\textsubscript{Value} {[sq mi]}\textsubscript{Unit}.''}), and one example is incorrect for multiple reasons (\emph{``90.60\% of the employed household members ({[17,257]}\textsubscript{Value}{[) work]}\textsubscript{Unit} within the municipality of {[Libon]}\textsubscript{Entity}. (implicit property: number of households)''}).

\paragraph*{Limitations.}
Wiki-Quantities has several limitations. Although the \{\{convert\}\} template allows for diverse expressions of quantities, the variety is limited. In particular, the dataset does not include annotations of quantities with percentages as the unit of measurement (Wiki-Measurements does). Template calls that refer to Wikidata are ignored. Finally, common units such as kilogram, meter, etc. are overrepresented (see Table~\ref{tab:wiki_quantities_unit_freqs}).

The latter limitation also applies to Wiki-Measurements where certain entity classes, and hence certain properties and units, are over-represented (see Tables~\ref{tab:wiki_measurements_freq_deduplicated} and \ref{tab:wiki_measurements_freq_balanced}). Additionally, the use of bots in the creation of Wikipedia articles may lead to similar sentences about the same entity class (e.g., different cities, rivers, or mountains). These are removed in Wiki-Quantities, but retained in Wiki-Measurements. Also, unlike Wiki-Quantities, Wiki-Measurements does not support compound quantities (e.g., 2 ft 1 in). Finally, qualifiers and quantity modifiers are considered optional and are only sporadically annotated. Unlike many other datasets, Wiki-Quantities does not include quantity modifiers in the quantity span annotations.

For each property, the quantitative statements were queried separately. To avoid hitting the timeout of the public Wikidata endpoint, a limit of 250\,000 results is set. Thus, not all quantitative statements of Wikidata are obtained. This can be circumvented by not querying the public endpoint, but hosting a local copy, or writing a parser for the database dump. However, only few queries run into the timeout, and not setting the limit would result in a more unbalanced dataset. 

Finally, parsing MediaWiki markup is imperfect. For example, many templates other than the \{\{convert\}\} template are not expanded, which may have introduced artifacts such as missing words or empty parentheses.

\section*{Usage Notes}

The easiest way to use the datasets unchanged for sequence labeling and question answering is to use the pre-processed versions for these tasks. If the use case is different or changes are desired, the raw datasets can be used as a starting point. To reproduce this work, the published code along with the accompanying Snakemake workflows can be used. Although the datasets are published in English, the methods are easily transferable to other languages. For those with more time and leisure, it is worth noting that the approach used to create Wiki-Quantities can be applied to scientific texts by using siunitx commands\footnote{\url{https://ctan.org/pkg/siunitx}} in LaTeX source files (e.g., from arXiv\footnote{\url{https://arxiv.org/}}) as annotations for quantities, numeric values, and units. When using or reproducing this work, please cite this data descriptor.

\section*{Code availability}
The code for creating the datasets is available under open licenses at:

\url{https://github.com/FZJ-IEK3-VSA/wiki-measurements}

\bibliography{bibliography}

\newpage
\section*{Acknowledgements}
The authors would like to thank the German Federal Government, the German State Governments, and the Joint Science Conference~(GWK) for their funding and support as part of the NFDI4Ing consortium. Funded by the German Research Foundation~(DFG) -- project number: 442146713. Furthermore, this work was supported by the Helmholtz Association under the program “Energy System Design”.

\section*{Author contributions statement}
J.G. conceived the methodology, created the datasets, and wrote the original draft. P.K., J.W., and D.S.  took care of the supervision, project administration, and funding acquisition. All authors reviewed the manuscript.

\section*{Competing interests}
The authors declare no competing interests.

\newpage
\section*{Figures \& Tables}

\begin{center}
{\small
\begin{lstlisting}[language=SPARQL, caption={SPARQL query used to get all quantitative properties in Wikidata.}, label=lst:property_query]
SELECT ?property ?propertyLabel
WHERE {  
    ?property wikibase:propertyType wikibase:Quantity .
    ?property rdfs:label|skos:altLabel ?propertyLabel .
    FILTER(LANG(?propertyLabel) = "en") .
}
\end{lstlisting}
}
\end{center}

\begin{center}
{\small
\begin{lstlisting}[language=SPARQL, caption={SPARQL query used to get up to \{limit\_threshold\} quantitative statements for a property \{property\_id\} in Wikidata.}, label=lst:quantitative_statement_query]
SELECT DISTINCT ?article ?entity ?value ?unit ?lowerbound ?upperbound ?qualifier ?qualifier_value ?qualifier_unit ?qualifier_lowerbound ?qualifier_upperbound ?qualifier_time_precision
WHERE {{  
    # Fix property    
    VALUES (?property) {{(wd:{property_id})}} 
    
    # Get associated wikipedia article   
    ?article schema:about ?entity .        
    ?article schema:isPartOf <https://{wiki_lang}.wikipedia.org/> .

    # Consider statements which have a quantity as the object    
    ?entity    ?p  ?statement .
    ?statement ?ps ?valuenode .
    ?property  wikibase:claim          ?p ;
            wikibase:statementValue ?ps .
    ?valuenode wikibase:quantityAmount ?value ;
            wikibase:quantityUnit   ?unit .    

    # Get upper and lower bound, if applicable
    OPTIONAL {{     
    ?valuenode wikibase:quantityLowerBound ?lowerbound ;
                wikibase:quantityUpperBound ?upperbound .
    }}
        
    # Get qualifiers, if applicable
    OPTIONAL {{    
        # Just some qualifier
        ?statement ?qualifier ?qualifier_value .    
        ?wdpq wikibase:qualifier ?qualifier . 

        OPTIONAL {{
            # A quantitative qualifier
            ?statement ?pqv ?pqv_ .
            ?wdpq wikibase:qualifierValue ?pqv .
            OPTIONAL {{             
                ?pqv_ wikibase:quantityUnit ?qualifier_unit .
            }}
            OPTIONAL {{
                ?pqv_ wikibase:quantityLowerBound ?qualifier_lowerbound ;
                    wikibase:quantityUpperBound ?qualifier_upperbound .     
            }}                    
            OPTIONAL {{                          
                ?pqv_ wikibase:timePrecision ?qualifier_time_precision .
            }}
            # (No precision for coordinate values, since it is mostly refered to locations in text by name)
        }}    
    }}      
}}        
LIMIT {limit_threshold}
\end{lstlisting}
}
\end{center}
\begin{table}[ht]
    \centering
    \caption{Dataset statistics for different variants of Wiki-Quantities}
    \begin{tabular}{lrrrr}
    \toprule     
      & Wiki-Quantities (raw) & Wiki-Quantities (large) &  Wiki-Quantities (small) &  Wiki-Quantities (tiny) \\
    \midrule
    \#Quantities        & 1\,227\,626 & 779\,943 & 59\,504 & 26\,076 \\
    \#Examples          & 832\,660    & 561\,393 & 44\,729 & 20\,148 \\
    \#Curated Examples  & 289         & 289      & 289     & 289     \\
    \#Wikipedia Pages   & 434\,860    & 289\,217 & 35\,758 & 17\,006 \\
    \bottomrule
    \end{tabular}
    \label{tab:wiki_quantities_stats}
\end{table}

\begin{table}[ht]
    \centering
    \caption{Comparing the English and Simple English Wikipedia as sources for Wiki-Quantities and Wiki-Measurements}
    \begin{tabular}{llrrr}
        \toprule
        Dataset & Wikipedia Language & \#Quantities  & \#Examples &\#Wikipedia Pages \\
        \midrule
        Wiki-Quantities (raw) & English & 1\,214\,927 & 822\,700  & 428\,941  \\
        & Simple English & 12\,699 & 9\,960 & 5\,919  \\ 
        \midrule
        Wiki-Measurements (raw) & English & 37\,790 & 37\,790 & 31\,131  \\
        & Simple English & 1\,659 & 1\,659 & 1\,327 \\
        \bottomrule
        \end{tabular}
        \label{tab:wiki_quantities_language_stats}
\end{table}

\begin{table}[ht]
    \caption{Top 50 most frequent units for different Wiki-Quantities variants. The percentage of all units in the respective dataset is given in parentheses next to the surface form of the unit. As to be expected, Wiki-Quantities (small) and (tiny) are more balanced with respect to units. Unit counts are based on rule-based parsing of the last quantity in a quantity span.}
    \centering
    \begin{tabular}{lllll}
\toprule
 & Wiki-Quantities (raw) & Wiki-Quantities (large) & Wiki-Quantities (small) & Wiki-Quantities (tiny) \\
\midrule
1 & mi (12.45) & m (9.68) & kW (3.27) & kW (2.19) \\
2 & km (10.94) & km (9.53) & km/h (2.73) & km (1.85) \\
3 & m (7.93) & ft (9.05) & km (2.69) & km/h (1.7) \\
4 & ft (7.65) & mi (8.14) & hp (2.66) & PS (1.65) \\
5 & in (6.67) & in (6.47) & m (2.58) & mi (1.63) \\
6 & kilometres (5.97) & miles (4.81) & °C (2.57) & hp (1.6) \\
7 & miles (3.97) & metres (4.23) & kg (2.52) & m (1.58) \\
8 & metres (3.47) & feet (4.19) & acres (2.49) & mph (1.53) \\
9 & feet (3.32) & kilometres (4.09) & lb (2.48) & kg (1.49) \\
10 & mm (3.27) & mm (3.24) & mph (2.46) & °C (1.47) \\
11 & acres (2.2) & acres (2.93) & °F (2.4) & acres (1.47) \\
12 & km2 (2.07) & cm (2.36) & ft (2.34) & ft (1.46) \\
13 & cm (1.96) & km2 (1.96) & mm (2.32) & °F (1.46) \\
14 & sq mi (1.73) & ha (1.7) & mi (2.28) & g (1.45) \\
15 & km/h (1.62) & °F (1.67) & ha (2.21) & t (1.45) \\
16 & °F (1.46) & °C (1.65) & cm (2.21) & L (1.44) \\
17 & mph (1.45) & km/h (1.47) & m2 (2.2) & lb (1.43) \\
18 & °C (1.44) & kg (1.29) & km2 (2.16) & in (1.38) \\
19 & ha (1.24) & mph (1.29) & in (2.16) & m3 (1.38) \\
20 & kg (1.19) & m2 (1.23) & metres (2.09) & mm (1.36) \\
21 & lb (1.05) & sq mi (1.18) & kilometres (2.07) & pounds (1.33) \\
22 & m2 (0.86) & lb (1.17) & miles (2.06) & kilometres (1.33) \\
23 & knots (0.76) & inches (0.9) & feet (2.05) & kilograms (1.31) \\
24 & inches (0.69) & hectares (0.68) & inches (1.91) & feet (1.31) \\
25 & centimetres (0.66) & centimetres (0.67) & long tons (1.71) & cm (1.31) \\
26 & millimetres (0.66) & millimetres (0.66) & t (1.5) & ha (1.31) \\
27 & long tons (0.64) & square kilometres (0.5) & pounds (1.43) & m2 (1.31) \\
28 & square kilometres (0.59) & kW (0.48) & PS (1.42) & millimetres (1.31) \\
29 & hectares (0.57) & meters (0.44) & hectares (1.18) & miles (1.3) \\
30 & nautical miles (0.54) & sq ft (0.44) & m3 (1.04) & km2 (1.3) \\
31 & meters (0.49) & nautical miles (0.41) & short tons (1.04) & hectares (1.28) \\
32 & square miles (0.44) & pounds (0.4) & g (0.92) & centimetres (1.26) \\
33 & kW (0.44) & hp (0.35) & millimetres (0.89) & metres (1.26) \\
34 & t (0.42) & square miles (0.35) & yards (0.87) & psi (1.17) \\
35 & pounds (0.34) & long tons (0.34) & kilograms (0.83) & yd (1.14) \\
36 & sq ft (0.33) & square feet (0.33) & centimetres (0.83) & meters (1.14) \\
37 & kilometers (0.3) & knots (0.32) & L (0.78) & knots (1.13) \\
38 & hp (0.27) & kilometers (0.32) & N$\cdot$m (0.7) & inches (1.12) \\
39 & g (0.24) & yards (0.3) & knots (0.69) & yards (1.1) \\
40 & square feet (0.23) & g (0.3) & miles per hour (0.68) & long tons (1.1) \\
41 & tonnes (0.22) & t (0.29) & tonnes (0.66) & tonnes (1.09) \\
42 & oz (0.21) & mile (0.27) & yd (0.65) & oz (1.01) \\
43 & yards (0.2) & kilograms (0.24) & oz (0.56) & cc (0.98) \\
44 & mile (0.2) & oz (0.24) & lb$\cdot$ft (0.56) & m3/d (0.98) \\
45 & kilograms (0.19) & m3 (0.24) & US gal (0.53) & miles per hour (0.97) \\
46 & m3 (0.18) & yd (0.22) & psi (0.53) & bhp (0.92) \\
47 & short tons (0.17) & short tons (0.22) & meters (0.52) & US gal (0.9) \\
48 & square kilometers (0.16) & miles per hour (0.21) & cc (0.51) & imp gal (0.82) \\
49 & yd (0.16) & nmi (0.18) & bhp (0.49) & lbf (0.81) \\
50 & nmi (0.16) & square metres (0.15) & imp gal (0.48) & l (0.8) \\
\bottomrule
\end{tabular}
    \label{tab:wiki_quantities_unit_freqs}
\end{table}

\begin{table}[ht]
\caption{Dataset statistics for different variants of Wiki-Measurements. Variants with added context are not included because they have the same statistics as their base variant.}
\centering
\begin{tabular}{llrrrr}
& & \multicolumn{4}{c}{\textbf{Wiki-Measurements}} \\
            &                         & large & large-strict & small & small-strict \\
\midrule
            & \#Pages                           & 32144 & 22332 & 19401 & 12292 \\
\midrule
            & \#Total                           & 38738 & 26009 & 21854 & 13257 \\
Examples    & \#Curated                         & 237   & 197   & 237   & 197 \\
            & \#Spatio-temporal Scopes Curated  & 217   & 190   & 217   & 190 \\
            & \#Qualifiers Curated              & 214   & 188   & 214   & 188 \\
\midrule
            & \#Entities                        & 38758 & 26012 & 21873 & 13260 \\
       Main & \#Properties                      & 30346 & 25978 & 15050 & 13226 \\
Annotations & \#Implicit Properties             & 17675 & 8173  & 12498 & 5361 \\
            & \#Values                          & 38738 & 26009 & 21854 & 13257 \\
            & \#Units                           & 36301 & 23917 & 20429 & 12018 \\
\midrule
            & \#Quantity Modifiers              & 2941  & 1819  & 2216 & 1345 \\
            & \#Temporal Scopes               & 1875  & 1624  & 728  & 557 \\
            & \#Determination Methods           & 560   & 472   & 208  & 173 \\
            & \#References                      & 143   & 131   & 50   & 40 \\
Optional    & \#Locations                       & 100   & 86    & 98   & 85 \\
Annotations & \#Applies to Part                 & 65    & 34    & 56   & 30 \\
            & \#Scopes                          & 73    & 47    & 62   & 39 \\
            & \#Start Times                     & 24    & 8     & 24   & 8 \\
            & \#Criteria Used                   & 15    & 9     & 9    & 4 \\
            & \#End Times                       & 5     & 3     & 5    & 3 \\
            & \#Other Qualifiers                & 394   & 138   & 259  & 106 \\
\bottomrule
\end{tabular}
\label{tab:wiki_measurements_stats}
\end{table}

\begin{table}[ht]
     \caption{Top 50 most frequent entities, properties, values, units, and temporal scopes in Wiki-Measurements (large). For entities, properties and temporal scopes the case is ignored. The share in percent is given in parentheses.}
    \centering   
\begin{tabular}{rrrrrr}
\toprule
 & Entities & Properties & Value & Units & Temporal Scope \\
\midrule
1 & it (2.83) & length (17.42) & three (0.79) & km2 (12.24) & 2010 (46.24) \\
2 & its (0.42) & area (17.42) & 10 (0.66) & square miles (6.96) & 2011 (12.0) \\
3 & coin (0.11) & total area (11.95) & 20 (0.62) & km (6.94) & 2020 (8.48) \\
4 & washington\,town...(0.07) & number of episodes (3.88) & two (0.6) & m (5.52) & 2016 (3.47) \\
5 & union township (0.07) & population (3.67) & 13 (0.57) & mile (4.83) & 2018 (3.47) \\
6 & sata (0.06) & number of stories (2.93) & 12 (0.52) & sq mi (4.22) & 2017 (2.4) \\
7 & lincoln township (0.06) & elevation (2.73) & 30 (0.46) & metres (3.68) & 2019 (2.19) \\
8 & liberty township (0.06) & elevation\,above\,mean...(2.4) & four (0.44) & feet (2.89) & 2014 (1.65) \\
9 & coins (0.05) & number of seats (2.03) & 40 (0.44) & episodes (2.64) & 2015 (1.39) \\
10 & jackson township (0.05) & diameter (1.44) & one (0.43) & mi (2.6) & 2012 (0.85) \\
11 & grant township (0.05) & total length (1.4) & 26 (0.43) & \$ (2.53) & 2021 (0.8) \\
12 & note (0.05) & height (1.19) & 50 (0.42) & ft (2.35) & 2001 (0.8) \\
13 & richland township (0.04) & number of beds (1.02) & 15 (0.41) & square\,kilom...(2.04) & 2013 (0.69) \\
14 & jefferson\,township\,(0.04) & average elevation (0.96) & 100 (0.41) & kilometres (1.96) & 2006 (0.64) \\
15 & lent (0.04) & passengers (0.85) & 5 (0.41) & square\,kilom...(1.94) & 2007 (0.64) \\
16 & town (0.04) & drainage basin (0.76) & 60 (0.4) & story (1.82) & 2000 (0.59) \\
17 & center township (0.04) & land area (0.74) & 2 (0.39) & miles (1.56) & 2004 (0.43) \\
18 & notes (0.04) & inhabitants (0.72) & 16 (0.38) & seat (1.51) & 1897 (0.32) \\
19 & madison\,township\,(0.04) & capacity (0.71) & 14 (0.36) & acts (1.1) & 1999 (0.32) \\
20 & three-dollar piece (0.04) & elevation above sea... (0.69) & 32 (0.35) & kilometre (0.96) & 2008 (0.27) \\
21 & city (0.03) & absolute magnitude (0.68) & 25 (0.34) & meters (0.95) & 2005 (0.27) \\
22 & water (0.03) & number\,of\,households\,(0.64) & 3 (0.34) & acres (0.9) & 2002 (0.21) \\
23 & the amazon (0.03) & width (0.63) & 1 (0.32) & kg (0.84) & 1960 (0.21) \\
24 & franklin township (0.03) & budget (0.62) & 11 (0.32) & metre (0.83) & december 2019 (0.16) \\
25 & cedar township (0.03) & altitude (0.61) & 18 (0.31) & bed (0.8) & february\,13,\,1905\,(0.16) \\
26 & walnut township (0.03) & production budget (0.55) & 8 (0.3) & mm (0.77) & july 24, 1936 (0.16) \\
27 & kallichore (0.03) & total land area (0.53) & 22 (0.3) & people (0.77) & 2009 (0.16) \\
28 & independence (0.03) & size (0.44) & 24 (0.27) & years (0.76) & january\,1,\,2018\,(0.16) \\
29 & marion township (0.03) & distance (0.43) & 6 (0.26) & hectares (0.74) & 1992 (0.16) \\
30 & gatwick (0.03) & surface area (0.42) & 19 (0.25) & cm (0.67) & july 13, 1936 (0.16) \\
31 & mercury (0.03) & atomic number (0.41) & 5,000 (0.25) & households (0.66) & january 2020 (0.11) \\
32 & three-dollar\,pieces\,(0.03) & number\,of\,employees\,(0.38) & 4 (0.25) & kilometers (0.58) & 1970 (0.11) \\
33 & summit township (0.03) & prominence (0.38) & 52 (0.25) & inhabitants (0.56) & 1945 (0.11) \\
34 & furlong (0.03) & albedo (0.34) & 200 (0.25) & episode (0.56) & 1984 (0.11) \\
35 & fremont township (0.03) & box office (0.31) & 23 (0.25) & days (0.53) & 1814 (0.11) \\
36 & triton (0.02) & episodes (0.31) & ten (0.23) & storey (0.53) & 1989 (0.11) \\
37 & garfield township (0.02) & weight (0.3) & 17 (0.23) & minutes (0.53) & 1 january 2018 (0.11) \\
38 & 2 bill (0.02) & seats (0.29) & 75 (0.22) & passengers (0.53) & january 2018 (0.11) \\
39 & the moon (0.02) & beam (0.28) & 28 (0.22) & MW (0.48) & july 22, 1926 (0.11) \\
40 & logan township (0.02) & total population (0.26) & 500 (0.22) & minute (0.45) & december\,31,\,20...(0.11) \\
41 & newfoundland (0.02) & orbital period (0.26) & five (0.22) & ha (0.39) & march 2014 (0.11) \\
42 & she (0.02) & stories (0.25) & 35 (0.22) & mg (0.39) & july 14, 1936 (0.11) \\
43 & acre (0.02) & mass (0.25) & 120 (0.22) & °C (0.35) & august 2017 (0.11) \\
44 & eden township (0.02) & beds (0.24) & 36 (0.21) & tracks (0.35) & 31\,december\,19...(0.11) \\
45 & hurdles (0.02) & beats per minute (0.22) & 21 (0.21) & ° (0.32) & year 2021 (0.11) \\
46 & germany (0.02) & inclination (0.21) & 45 (0.21) & employees (0.31) & july 2020 (0.11) \\
47 & finasteride (0.02) & altitude\,above\,sea\,l...\,(0.21) & 300 (0.21) & acre (0.31) & june 29, 1994 (0.11) \\
48 & long island (0.02) & seating capacity (0.2) & 39 (0.2) & act (0.29) & 1994 (0.11) \\
49 & clay township (0.02) & basin area (0.15) & 33 (0.2) & hours (0.28) & july 10, 1913 (0.11) \\
50 & fairview\,township\,(0.02) & geographic area (0.14) & 150 (0.2) & mg/m3 (0.26) &  for\,the\,2000 mil..(0.05) \\
\bottomrule
\end{tabular}
    \label{tab:wiki_measurements_freq_deduplicated}
\end{table}

\begin{table}[ht]
     \caption{Top 50 most frequent entities, properties, values, units, and temporal scopes in Wiki-Measurements (small). For entities, properties and temporal scopes the case is ignored. The share in percent is given in parentheses.}
    \centering   
\begin{tabular}{rrrrrr}
\toprule
 & Entities & Properties & Value & Units & Temporal Scope \\
\midrule
1 & it (1.43) & length (20.05) & three (1.08) & km (7.4) & 2010 (16.48) \\
2 & its (0.27) & area (9.55) & 10 (0.91) & km2 (5.85) & 2011 (11.95) \\
3 & coin (0.18) & number of episodes (6.9) & 20 (0.88) & m (5.66) & 2020 (6.87) \\
4 & coins (0.09) & number of stories (5.06) & two (0.86) & mile (4.38) & 2017 (5.49) \\
5 & note (0.08) & population (3.12) & 13 (0.83) & episodes (4.09) & 2018 (5.08) \\
6 & lent (0.08) & number of seats (2.98) & 12 (0.75) & metres (3.83) & 2019 (4.4) \\
7 & sata (0.07) & elevation (2.47) & four (0.69) & \$ (3.02) & 2016 (3.71) \\
8 & notes (0.07) & total area (2.46) & one (0.67) & story (2.74) & 2014 (3.43) \\
9 & three-dollar\,piece\,(0.06) & number of beds (1.75) & 40 (0.65) & ft (2.55) & 2015 (2.88) \\
10 & the amazon (0.05) & height (1.61) & 30 (0.61) & sq mi (2.45) & 2021 (1.79) \\
11 & three-dollar\,pieces\,(0.05) & passengers (1.45) & 26 (0.59) & mi (2.35) & 2001 (1.65) \\
12 & triton (0.04) & diameter (1.36) & 50 (0.59) & kilometres (2.16) & 2006 (1.51) \\
13 & 2 bill (0.04) & capacity (1.28) & 60 (0.58) & miles (2.01) & 2012 (1.37) \\
14 & water (0.04) & inhabitants (1.0) & 100 (0.56) & seat (1.9) & 2000 (1.24) \\
15 & the moon (0.04) & land area (0.95) & 15 (0.56) & square\,kilom...(1.84) & 2007 (1.24) \\
16 & she (0.04) & budget (0.86) & 16 (0.54) & feet (1.7) & 2004 (0.96) \\
17 & hurdles (0.04) & altitude (0.86) & 2 (0.5) & kilometre (1.45) & 2013 (0.69) \\
18 & 5 note (0.03) & total length (0.8) & 14 (0.49) & square miles (1.41) & 1999 (0.69) \\
19 & gatwick (0.03) & distance (0.75) & 25 (0.47) & acts (1.4) & 1897 (0.55) \\
20 & finasteride (0.03) & elevation\,above\,sea...\,(0.71) & 5 (0.47) & bed (1.19) & december 2019 (0.41) \\
21 & arcturus (0.03) & number\,of\,employees\,(0.68) & 1 (0.46) & meters (1.18) & 2008 (0.41) \\
22 & mercury (0.03) & total land area (0.66) & 11 (0.46) & mm (1.15) & 2002 (0.41) \\
23 & day (0.03) & size (0.63) & 32 (0.44) & years (1.13) & 2005 (0.41) \\
24 & acre (0.03) & width (0.59) & 18 (0.44) & metre (1.07) & 1992 (0.41) \\
25 & telescope (0.03) & episodes (0.56) & 3 (0.43) & kg (1.05) & january 2020 (0.27) \\
26 & aes (0.03) & seats (0.55) & 22 (0.41) & cm (1.01) & 1970 (0.27) \\
27 & shvak (0.03) & box office (0.52) & 8 (0.38) & hectares (0.93) & 1945 (0.27) \\
28 & 10 bill (0.03) & weight (0.5) & 24 (0.38) & acres (0.91) & 1984 (0.27) \\
29 & city (0.02) & drainage basin (0.5) & 500 (0.36) & episode (0.91) & 1989 (0.27) \\
30 & fast (0.02) & stories (0.45) & 52 (0.36) & minutes (0.88) & 1 january 2018 (0.27) \\
31 & frankfurt airport (0.02) & number\,of\,households\,(0.45) & 19 (0.36) & storey (0.81) & july 24, 1936 (0.27) \\
32 & 6502 (0.02) & beds (0.43) & ten (0.35) & passengers (0.79) & january 2018 (0.27) \\
33 & rack (0.02) & surface area (0.37) & five (0.33) & days (0.73) & 2009 (0.27) \\
34 & newfoundland (0.02) & mass (0.37) & 200 (0.33) & minute (0.72) & december\,31,\,20...(0.27) \\
35 & norethisterone (0.02) & prominence (0.37) & 21 (0.32) & people (0.71) & march 2014 (0.27) \\
36 & hanukkah (0.02) & absolute magnitude (0.35) & 28 (0.32) & MW (0.7) & july 2020 (0.27) \\
37 & pakistan (0.02) & seating capacity (0.35) & 23 (0.32) & mg (0.68) & 1960 (0.27) \\
38 & howitzer m1 (0.02) & beam (0.32) & 300 (0.31) & inhabitants (0.67) & 1994 (0.27) \\
39 & germany (0.02) & orbital period (0.29) & 35 (0.31) & square\,kilom...(0.62) & july 13, 1936 (0.27) \\
40 & sedna (0.02) & total population (0.28) & 75 (0.3) & tracks (0.61) & july 10, 1913 (0.27) \\
41 & beta pictoris (0.02) & altitude\,above\,sea\,l... (0.27) & 39 (0.3) & ha (0.52) & for\,the\,2000\,mil..(0.14) \\
42 & furlong (0.02) & apparent magnitude (0.25) & 5,000 (0.29) & kilometers (0.51) & erected in 1979 (0.14) \\
43 & raloxifene (0.02) & beats per minute (0.25) & 17 (0.29) & employees (0.49) & current (0.14) \\
44 & everest (0.02) & cost (0.24) & 45 (0.29) & acre (0.49) & in\,january\,2012...(0.14) \\
45 & chain (0.02) & staff (0.24) & 120 (0.27) & °C (0.49) & now (0.14) \\
46 & highway 1 (0.02) & half-life (0.23) & 70 (0.27) & hours (0.45) & around 2005 (0.14) \\
47 & titanic (0.02) & average elevation (0.21) & 6 (0.27) & households (0.41) & on\,october\,29,...(0.14) \\
48 & today (0.02) & running time (0.21) & 4 (0.27) & act (0.41) & officially\,open...(0.14) \\
49 & alabama (0.02) & injured (0.21) & twelve (0.26) & meter (0.35) & started\,in\,2009...(0.14) \\
50 & naltrexone (0.02) & life expectancy (0.21) & 29 (0.26) & hour (0.34) & finished\,in\,late...(0.14) \\
\bottomrule
\end{tabular}
    \label{tab:wiki_measurements_freq_balanced}
\end{table}

\end{document}